\pdfoutput=1

\documentclass[11pt]{article}

\usepackage[final]{acl}

\usepackage{times}
\usepackage{latexsym}
\usepackage{subcaption}
\usepackage{booktabs}
\usepackage{adjustbox}
\usepackage{graphicx}
\usepackage{amsmath}
\usepackage{tcolorbox}
\usepackage{multirow}
\usepackage[T1]{fontenc}

\usepackage[utf8]{inputenc}

\usepackage{microtype}

\usepackage{inconsolata}

\usepackage{graphicx}

\newcommand{\aaf}{\vspace*{-6pt}}

\title{Impact of Fine-Tuning Methods on Memorization\\in Large Language Models}

\author{
  Jie Hou \quad
  Chuxiong Wu \quad
  Lannan Luo \quad
  Qiang Zeng \\
  George Mason University \\
  \texttt{\{jhou4, cwu27, lluo4, zeng\}@gmu.edu}
}

\begin{document}

\maketitle

\begin{abstract}
As the capabilities of pre-trained large language models (LLMs) continue to advance, the ``pre-train and fine-tune'' paradigm has become increasingly mainstream, leading to the development of various fine-tuning methods. However, the privacy risks arising from memorization during fine-tuning have received relatively little attention. To address this gap, we categorize popular fine-tuning approaches and assess their impact on memorization through the lens of
membership inference attacks (MIAs). Our results show that, compared to \emph{parameter-based} fine-tuning, \emph{prompt-based} fine-tuning achieves competitive performance while exhibiting lower vulnerability to MIAs. Furthermore, prompt-based methods maintain low memorization regardless of model scale. These findings suggest that parameter-based fine-tuning is more prone to leaking private information, whereas prompt-based fine-tuning serves as a more privacy-preserving option.

\end{abstract}

\section{Introduction}\label{sec:1}

The training of a large language model (LLM) typically involves two stages: pre-training and fine-tuning~\cite{dong2019unified,radford2018improving}. In the pre-training stage, models are trained on large, general-domain corpora to capture broad linguistic patterns and representations. Subsequently, during fine-tuning, the pre-trained models are adapted to specific downstream tasks using task-specific datasets.

Extensive research~\cite{carlini2022membership,shi2023detecting,mattern-etal-2023-membership,meeus2024did,wuyou,yeom2018privacy} has demonstrated that LLMs can memorize training data during the pre-training stage, potentially leading to data leakage risks. 
While existing studies~\cite{mattern-etal-2023-membership,shi2023detecting,meeus2024did,wuyou} have primarily examined memorization during pre-training, there remains a significant gap in understanding how fine-tuning influences the retention and exposure of fine-tuning data. Addressing this gap is important because fine-tuning involves a diverse range of methods that vary substantially in parameter updates, structural modifications, and optimization strategies, potentially leading to different levels of memorization. 
To address this gap, the key research question we explore is: \textbf{How do different fine-tuning methods impact the model’s memorization of fine-tuning data?}

Fine-tuning methods can be broadly categorized into two types. 
(1) \textbf{Parameter-based fine-tuning} updates model parameters, either fully or via lightweight, trainable components such as Low-Rank Adaptation (LoRA)~\cite{hu2021lora}. 
(2) \textbf{Prompt-based fine-tuning} keeps the pre-trained model parameters frozen and instead learns task-specific prompts (e.g., soft tokens or prefix embeddings). 
Our study includes five representative methods spanning both categories.
To evaluate the memorization effects of fine-tuning, we use \textit{membership inference attacks} (MIAs), which aim to determine whether a specific data point was part of the training set—thereby serving as an indicator of memorization. 
A wide range of MIAs are employed in our experiments~\cite{yeom2018privacy,mireshghallah2022quantifying,carlini2022membership,shi2023detecting}.

Our experiments utilize three widely used datasets: Wikitext~\cite{merity2016pointer} for general language model fine-tuning, WebNLG~\cite{gardent2017creating} for structured data-to-text generation, and Xsum~\cite{xsum} for abstractive summarization. Together, they simulate realistic scenarios where pre-training is followed by task-specific adaptation.
Our results yield a series of important and insightful findings. 
We further provide interpretations to explain key observations.

We make the following contributions:

\begin{itemize}
    \item To the best of our knowledge, this is the first study to systematically investigate the memorization behaviors of LLMs across a range of fine-tuning methods.

    \item Our study uncovers several important findings. For example, 
    compared to \emph{parameter-based} fine-tuning, \emph{prompt-based} fine-tuning achieves competitive model performance while exhibiting lower vulnerability to MIAs. Moreover, increasing model size significantly amplifies memorization in parameter-based fine-tuning, whereas prompt-based fine-tuning remains largely insensitive to model scale and maintains consistently low memorization levels. These results suggest that parameter-based fine-tuning is more prone to leaking private information, while prompt-based fine-tuning offers a more privacy-preserving option.
    
    \item We provide interpretations to support and explain key observations.
\end{itemize}

\textbf{Roadmap.} Section \ref{sec:2} provides an overview of fine-tuning methods discussing their key characteristics. Section \ref{sec:3} introduces the concept of memorization in LLMs and presents MIA methods used to quantify memorization. Section \ref{sec:4} describes the experimental setup, including datasets, model configurations, and evaluation metrics. Section \ref{sec:5} presents the experimental results, followed by an in-depth analysis and discussion of memorization across different fine-tuning paradigms. Section \ref{sec:6} discusses the impact of pre-trained models and downstream tasks on memorization across different fine-tuning paradigms. Finally, Section \ref{sec:7} concludes the paper and outlines potential directions for future research.

\section{Fine-Tuning Methods}\label{sec:2}

\textbf{Parameter-based fine-tuning} updates model parameters, either directly or through lightweight trainable components. Full Fine-Tuning (FFT) is rarely used for LLMs because it updates all parameters, making it computationally expensive and unsuitable for multi-task settings, where task-specific adaptations should not overwrite pre-trained knowledge.
We thus examine two representative methods in this category:
  \begin{itemize}
    \item \textbf{Model Head Tuning (FT head)}: fine-tunes only the final output layer (e.g., classification head), keeping all other pre-trained model parameters frozen.
    \item \textbf{Low-Rank Adaptation (LoRA)}~\cite{hu2021lora}: introduces trainable low-rank matrices into the attention layers, enabling efficient fine-tuning while freezing the original model weights.

  \end{itemize}

  \textbf{Prompt-based fine-tuning} freezes model parameters and instead learns task-specific prompts. This study considers three representative prompt-based techniques:
\begin{itemize}
    \item \textbf{Prefix Tuning}~\cite{li2021prefix}: prepends trainable prefix vectors to the keys and values of each attention layer.
    \item \textbf{Prompt Tuning}~\cite{lester2021power}: optimizes a small set of continuous prompt embeddings that are directly prepended to the model input embeddings.
    \item \textbf{P-tuning}~\cite{liu2021gpt}: generates trainable continuous prompts via an additional neural network %
    that are then incorporated into the input to better guide the model.

\end{itemize}

In comparison, \citet{mireshghallah2022empirical} examines only parameter-based fine-tuning methods and focuses on unstructured datasets.

\section{Memorization and MIAs}\label{sec:3}
Modern deep learning models learn mappings between inputs and outputs through large-scale data training. However, beyond extracting generalizable patterns, models may also memorize individual training samples due to overfitting~\cite{yeom2018privacy}, a phenomenon known as \emph{model memorization}. 
MIAs~\cite{yeom2018privacy,mattern-etal-2023-membership,mireshghallah2022quantifying,shi2023detecting} have been widely adopted to study model memorization. In this work, we quantify the impact of fine-tuning methods on model memorization
of fine-tuning data using MIAs. The following MIA methods are employed in our study:

\noindent\textbf{LOSS:} This method~\cite{yeom2018privacy} uses the sample's loss as its membership score, 
$$Score = L( x , M_t)$$
where $L(x, M_t)$ denotes the loss of the target model $M_t$ on input $x$.

\noindent\textbf{Reference-based method (Ref):} \citet{mireshghallah2022quantifying} extended the ideas proposed by \citet{shokri2017membership} and \citet{carlini2019secret} by utilizing a reference model to calibrate the loss. Specifically, for a target model $M_t$, a pre-trained model without fine-tuning is used as the reference model $M_r$. The membership score for a sample \( x \) is then defined as the difference between the losses of \( x \) on \( M_t \) and \( M_r \).
$$Score = L( x , M_t) - L( x , M_r)$$

\noindent\textbf{Zlib Entropy (Zlib):} \citet{carlini2021extracting} proposed using the ratio of the sample's loss to its zlib entropy as the membership score.
$$Score = \frac{L( x , M_t)}{zlib(x)}$$
\noindent\textbf{Min-K\%:} \citet{shi2023detecting} proposed using the average log-likelihood of the lowest \( k\% \) token probabilities in the text as the membership score.
$$Score = \frac{1}{E}\sum_{x_i\in Min-K\%(x)}\log p(x_i|x_1,x_2...,x_{i-1})$$
where \( x = x_1, x_2, \ldots, x_N \) is the input token sequence, \( p(x_i|x_1, \ldots, x_{i-1}) \) is the model's predicted probability for token \( x_i \) given its preceding context, and $Min-K\%(x)$ denotes the set of tokens with the lowest $k\%$ predicted probabilities in the sequence. \( E \) is the number of tokens in this set.

\section{Experimental Setup}\label{sec:4}
\subsection{Datasets}\label{sec:4.1} We conducted experiments on three publicly available datasets across different downstream tasks: (1) Wikitext-2-raw-1 (\texttt{Wikitext})~\cite{merity2016pointer} is a dataset derived from Wikipedia. It preserves the original structure and formatting of Wikipedia text and includes approximately two million words. (2) \texttt{WebNLG}~\cite{gardent2017creating} is specifically designed for natural language generation tasks. It comprises 35.4k training samples, each consisting of a structured triple (Subject-Predicate-Object) as input and a coherent sentence as output, spanning a variety of domains. (3) \texttt{Xsum}~\cite{xsum} is a dataset created for the task of abstractive summarization. It consists of BBC news articles paired with a single-sentence summary that captures the key point of each article. Among them, Wikitext is an unstructured text dataset, while WebNLG and Xsum are structured datasets, designed for data-to-text generation and abstractive summarization, respectively.

To prepare the datasets for fine-tuning, we follow standard preprocessing strategies tailored to each benchmark. On the Wikitext dataset, we adopt the methodology proposed by \citet{mireshghallah2022quantifying}, segmenting the text into blocks of 128 tokens. On the WebNLG dataset, we employ the preprocessing strategy outlined in Prefix Tuning~\cite{li2021prefix}. 
Given the large size of the Xsum dataset, we use a subset of the training set by randomly selecting 5,000 samples for fine-tuning.

\begin{table}
\begin{tabular}{llll}
\hline
Dataset  & Fine-tuning & Mem  & Non-Mem \\ \hline
Wikitext & 22,335       & 2,000 & 2,000    \\
WebNLG   & 35,426       & 1,779 & 1,779   \\
Xsum   & 5,000       & 1,000 & 1,000   \\\hline
\end{tabular}
\caption{Data split stats for fine-tuning and MIAs}
\label{tab:dataset_stat}
\end{table}

To evaluate memorization, as summarized in Table~\ref{tab:dataset_stat}, we construct the MIA evaluation dataset by sampling an equal number of instances from the training and test sets, labeling them as membership and non-membership samples, respectively. Specifically, the test sets of the Wikitext and WebNLG datasets are used as non-membership samples, while an equal number of samples randomly drawn from the corresponding training sets serve as membership samples. For the Xsum dataset, we sample 1,000 instances from both the training and test sets to construct the membership and non-membership samples. We use the validation splits (not shown in Table~\ref{tab:dataset_stat}) of the three datasets to compute the validation perplexity (PPL).

\subsection{Target models} \label{sec:4.2}

Our study involves three representative open-source LLMs: LLaMA 2-7B~\cite{touvron2023llama}, GPT2-series~\cite{radford2019language}, and LLaMA 3-1B~\cite{touvron2023llama}.

\subsection{Evaluation Metrics}\label{sec:4.3} 
We utilize \emph{\textbf{validation PPL}} as the primary metric for evaluating model performance. To evaluate model memorization, we perform MIAs against  fine-tuned  models and report the area under the \emph{receiver operating characteristic curve (\textbf{AUC-ROC})} as the metric. 

\subsection{Implementation Details}\label{sec:4.4} 
We fine-tune the GPT2-series, LLaMA2-7B, and LLaMA3-1B models for 15 epochs each, which is sufficient for effective fine-tuning and for studying the impact of overfitting (Section~\ref{sec:5.3}). Optimization is performed using AdamW with a learning rate of 5e-5, and a linear scheduler is employed to stabilize training and improve convergence.

We adhere to the fine-tuning settings specified in the original papers. Specifically, for LoRA, we set the rank of the low-rank matrix $r$ to 16 and $\alpha$ to 32, with the adaptation applied to the query and key projection matrices, following the default setting. For Prefix Tuning, we follow \citet{li2021prefix} and use 5 virtual tokens. For P-tuning and Prompt Tuning, we adopt the configurations from \citet{liu2021gpt} and \citet{lester2021power}, using 20 and 8 virtual tokens, respectively.

We fine-tune the GPT2-series models on an NVIDIA RTX 4090, with each epoch taking approximately 25 minutes for smaller models and up to around 2 hours for the largest. LLaMA2-7B and LLaMA3-1B are fine-tuned on an NVIDIA H100, requiring about 2 hours per epoch.

\section{Results and Observations}\label{sec:5}

\begin{table*}[h]
\centering

\begin{tabular}{clcc|cccc}
\toprule
\multirow{2}{*}{\textbf{Dataset}} & \multirow{2}{*}{\textbf{Method}} & \multicolumn{2}{c|}{\textbf{Parameter-based}} & \multicolumn{3}{c}{\textbf{Prompt-based}} \\ \cline{3-7} & &
\textbf{FT head} & \textbf{LoRA} & \textbf{Prefix tuning} &
\textbf{Prompt tuning} & \textbf{P-tuning}   \\\midrule
\midrule
\multirow{5}{*}{WebNLG} 
& LOSS       & 0.9047 &  0.8516 & 0.6213 & 0.5816 & 0.6260   \\
& Ref        & 0.6779 &  0.6719 & 0.6009 & 0.5761 & 0.5902  \\
& Zlib       & 0.8566 & 0.8383 &  0.5516 & 0.5402 & 0.5889 \\
& Min-K\%    & 0.4616 & 0.4667 & 0.4616 & 0.5332 & 0.5264\\
\cline{1-7}
\multicolumn{2}{c}{Validation PPL}          & 1.5432  &  1.5198 &  1.9769 & 1.7198  &    1.6427  \\
\midrule
\multirow{5}{*}{Wikitext} 
& LOSS       & 0.8886 & 0.7865 & 0.4843 & 0.4783 & 0.4871 \\
& Ref        &  0.8754 & 0.7930 & 0.4956 & 0.4881 & 0.4976\\
& Zlib       & 0.9303 & 0.8405 & 0.4976 &  0.4867 & 0.4968\\
& Min-K\%    & 0.8723 & 0.7332 & 0.4931 & 0.5233 & 0.5233\\
\cline{1-7}
\multicolumn{2}{c}{Validation PPL}    & 9.0279  &  8.2709 &  11.0392 & 7.9588  &    7.5913  \\
\midrule
\multirow{5}{*}{Xsum} 
& LOSS       & 0.8231 & 0.7155 & 0.5224 & 0.5276 & 0.5318 \\
& Ref        & 0.7239 & 0.6454 & 0.4935 & 0.4935 & 0.5025\\
& Zlib       & 0.6728 & 0.5956 &  0.4994 & 0.5047 & 0.5076\\
& Min-K\%    & 0.9649 & 0.8297 & 0.4980 & 0.4876 & 0.4792\\
\cline{1-7}
\multicolumn{2}{c}{Validation PPL}    & 4.4514  &  4.5113 &  6.3430 & 4.2383  &    4.1559  \\

\bottomrule
\end{tabular}

\caption{Comparison of MIA performance (AUC), and model performance (validation PPL) across different fine-tuning methods. (A higher AUC indicates a greater memorization level, and a lower validation PPL suggests better task performance.) Results are reported for LLaMA2-7B after 5 epochs, when it is effectively fine-tuned without overfitting (see Figure~\ref{fig:PPL_results}).}
\label{tab:AUC}

\end{table*}

\subsection{Memorization across Tuning Methods}\label{sec:5.1}

We begin our analysis by examining the relationship between fine-tuning methods and memorization. Specifically, we ask: \textit{Does the choice of fine-tuning strategy affect how much a model memorizes its training data for fine tuning?}

We use LLaMA2-7B as a representative example; similar observations hold for the other models.
Table~\ref{tab:AUC} %
provides the results across the WebNLG, Wikitext, and Xsum datasets.

\begin{tcolorbox}[sharp corners, colback=white!30, colframe=black, title=\textbf{Observation $\sharp$1}]
Parameter-based fine-tuning demonstrates a higher tendency to explicitly memorize training data.
\end{tcolorbox}

In general, all fine-tuning methods achieve comparable task performance in terms of validation PPL, but prompt-based methods consistently exhibit significantly lower memorization than their parameter-based counterparts. As shown in Table~\ref{tab:AUC}, parameter-based methods such as FT head and LoRA exhibit high vulnerability to MIAs, with the highest AUC scores exceeding 0.8, indicating substantial memorization of training data. In contrast, prompt-based methods, including Prefix Tuning, Prompt Tuning, and P-tuning, achieve competitive task performance while demonstrating much lower memorization. For example, on WebNLG, P-tuning's validation PPL is only slightly higher than that of FT head, yet its AUC remains between 0.55 and 0.65, close to random guessing.

\begin{tcolorbox}[sharp corners, colback=white!30, colframe=black, title=\textbf{Observation $\sharp$2}]
Parameter-based fine-tuning exhibits increasing memorization over training epochs, while prompt-based fine-tuning maintains consistently low memorization throughout training.
\end{tcolorbox}

\begin{figure}
    \centering
    
        \includegraphics[width=\linewidth]{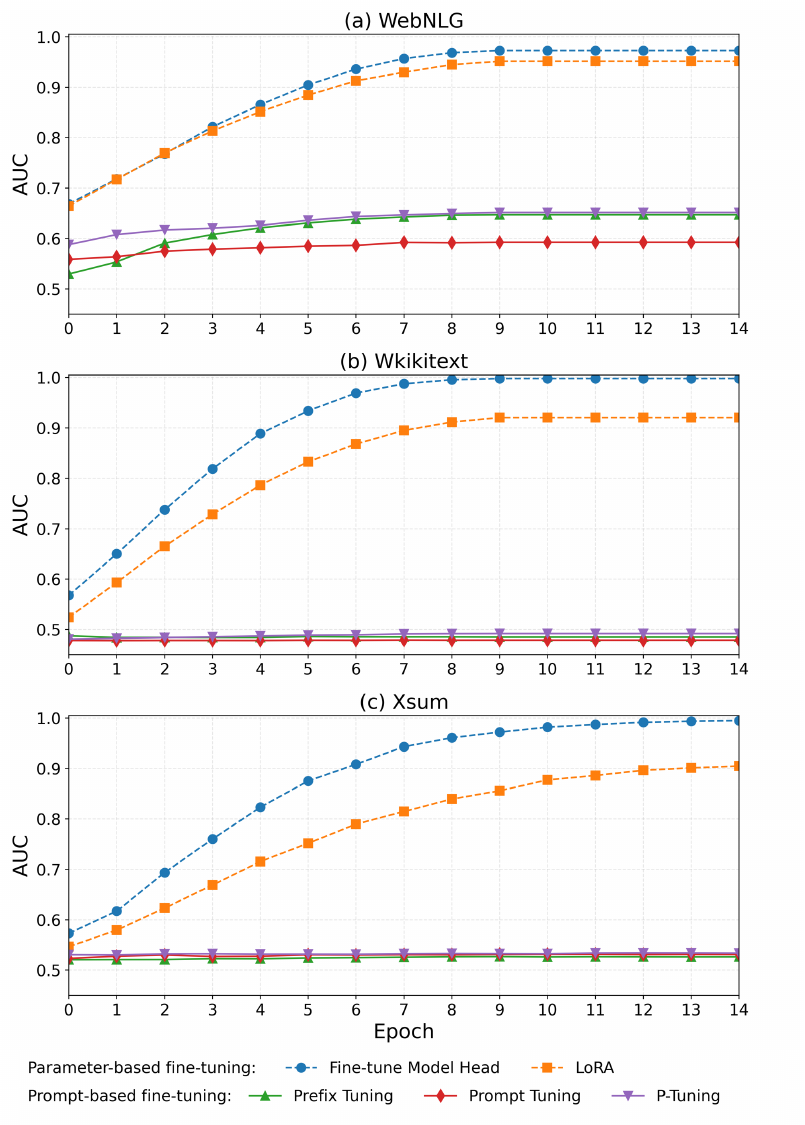}

    \aaf
    \caption{Performance of the LOSS attack against LLaMA2-7B on WebNLG, Wikitext, and Xsum.}
    \label{fig:MIA_results}
    \aaf   
\end{figure}

We further analyze how memorization evolves during training for different fine-tuning paradigms. As shown in Figure~\ref{fig:MIA_results}, parameter-based fine-tuning exhibits a consistent upward trend in AUC across training epochs, indicating that memorization intensifies as training progresses. By contrast, \emph{prompt-based fine-tuning shows no notable increase in memorization over time}, maintaining low AUC scores throughout training.

 \begin{figure}[h]
     \centering
     \includegraphics[width=\linewidth]{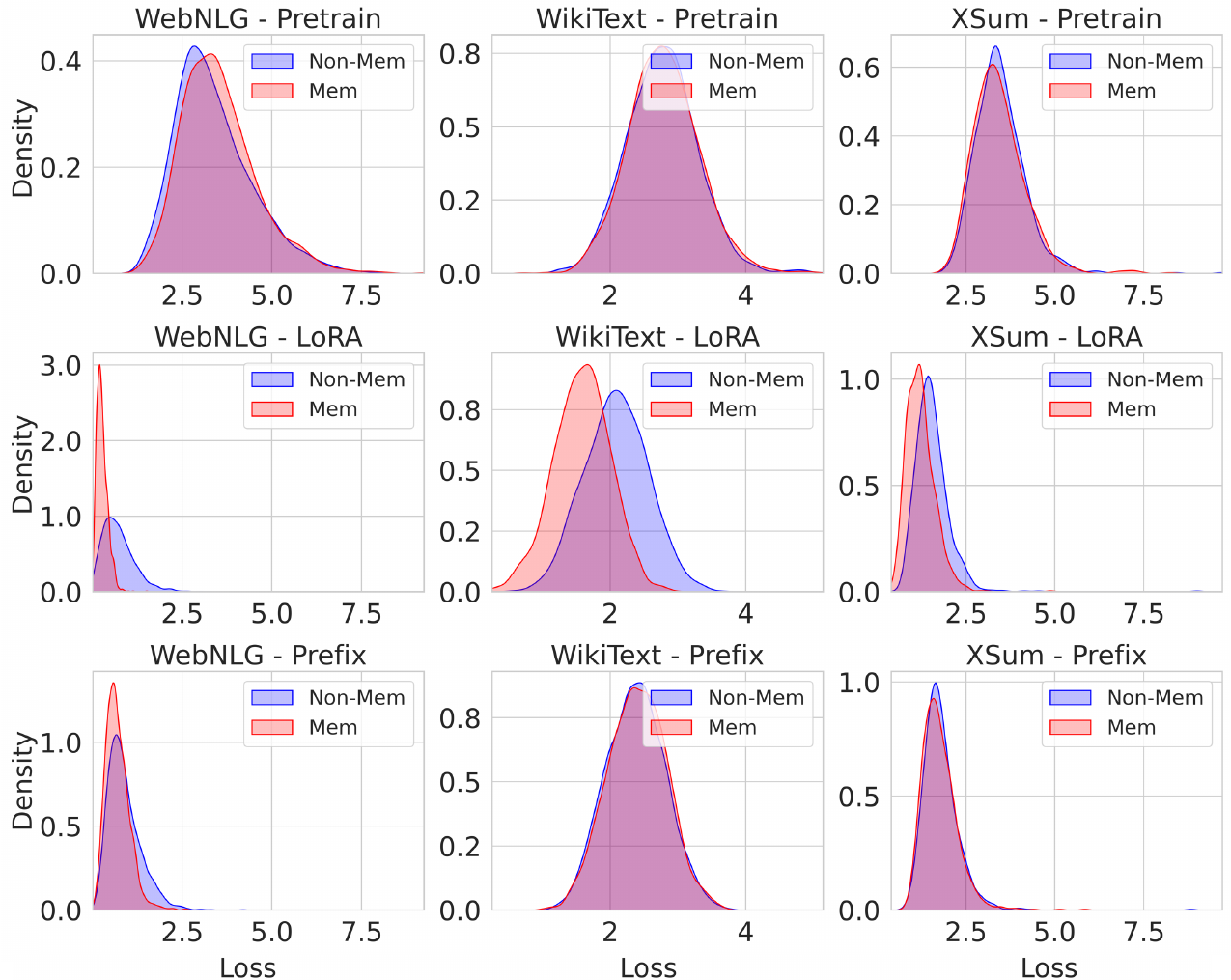}
     \caption{Loss distributions of membership and non-membership samples on LLaMA2-7B under three model settings: (a) pre-trained only, (b) fine-tuned with LoRA, and (c) fine-tuned with Prefix Tuning. }
     \label{fig:loss_distribution}
 \end{figure}

\subsection{Why Prompt-Based Fine-Tuning Exhibits Low Memorization}\label{sec:5.2}

The relatively low memorization observed in prompt-based tuning motivates a closer examination of its underlying mechanisms and training behavior.  One possible explanation for this result is that prompt-based fine-tuning introduces a bias into the model’s attention mechanism indirectly via the soft prompt or prefix, rather than altering the attention mechanism itself. This indirect influence may reduce the model’s susceptibility to MIAs. 

Taking a specific transformer block in prefix tuning as an example, for an input sequence ($t_1,t_2,...,t_{i-1}$), the output from the pre-trained model is obtained as follows:
$$t_i = \sum_{h=1}^{H} \sum_{j=1}^{p} A_{ij}^h W^h_V x_j$$
where \(t_i\) represents the output feature at position \(i\). The input sequence has a length of \(p\), corresponding to the number of tokens, and the computation involves \(H\) attention heads. \(A_{ij}^h\) denotes the attention weight from position \(i\) to position \(j\) in the \(h\)-th attention head. \(W^h_V\) is the value matrix associated with the \(h\)-th attention head. \(x_j\) refers to the input feature vector at position \(j\). 

As demonstrated by \citet{petrov2023prompting}, we can formally establish:
$$t_i^{pt} = A^{pt}_{i0}W_VS_1  + (1-A_{i0}^{pt})t_i$$
\citet{petrov2023prompting} prove that the presence of a prefix does not alter the relative distribution of the input but only shifts the attention to different content. Such a shift enables the fine-tuned model to solve similar tasks by leveraging the capabilities of the pre-trained model. However, for novel tasks like Wikitext, which demands modeling complex contextual language dependencies, prompt-based fine-tuning fails to learn new attention patterns, unlike for structured tasks such as WebNLG where input-output alignment is more explicit.

Meanwhile, to validate our hypothesis, we compute the distributions of non-membership and membership examples on the LLaMA2-7B model under three settings: the pre-trained model, the model fine-tuned with LoRA, and the model fine-tuned with prefix tuning. The results are shown in Figure \ref{fig:loss_distribution}. The experimental results indicate that after fine-tuning with LoRA, there is a significant difference in the distributions of membership and non-membership samples. However, after prefix tuning, the difference between these distributions is smaller, particularly on the Wikitext datasets. This is consistent with the theoretical explanation described above, where we discussed how parameter-based fine-tuning leads to overfitting and a distribution shift between training and unseen data. In contrast, prompt-based fine-tuning, which optimizes a soft prompt or prefix, does not induce such a representational shift; as a result, models fine-tuned via prompt-based methods demonstrate reduced memorization of the training data.

\subsection{Performance in Different Tuning Paradigms}\label{sec:5.3}

\begin{figure}[h]
    \centering
    \includegraphics[width=\linewidth]{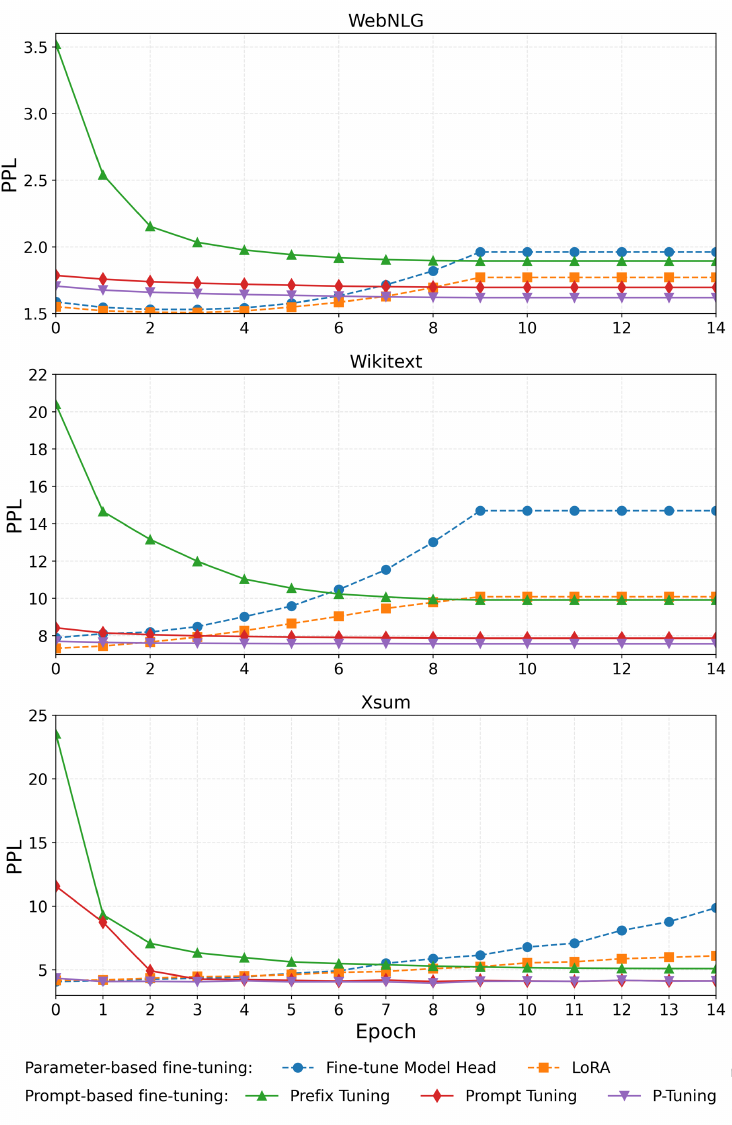}

    \caption{Validation PPL against LLaMA2-7B on WebNLG, Wikitext and Xsum. }
    \label{fig:PPL_results}
\end{figure}

We use validation PPL as the metric to evaluate the performance of different fine-tuning paradigms on downstream tasks with LLaMA2-7B.
Figure \ref{fig:PPL_results} presents the validation PPL over multiple training epochs. While both fine-tuning paradigms achieve similar optimal validation PPL (as shown in Table~\ref{tab:AUC}), they exhibit distinct learning trajectories that remain consistent across datasets. 
In parameter-based fine-tuning, the validation PPL initially decreases over the first few epochs but later increases due to overfitting, before eventually converging. In contrast, prompt-based fine-tuning maintains a slightly decreasing validation PPL throughout training, converging without the overfitting-induced rise observed in parameter-based methods. This phenomenon supports the explanation in Section \ref{sec:5.2}, which suggests that prompt-based fine-tuning does not significantly alter the internal sample distribution of the model, but instead introduces a bias that shifts the entire sample space toward one better suited for the downstream task.

\section{Discussion}\label{sec:6}

\begin{table*}
\centering

\scalebox{0.95}{\begin{tabular}{lccccccccc}
\toprule
\multirow{2}{*}{\textbf{Pre-trained model}} & \multirow{2}{*}{\textbf{\#Params}} & \multicolumn{3}{c|}{\textbf{Parameter-based}} & \multicolumn{3}{c}{\textbf{Prompt-based}} \\ \cline{3-8}  & &
\textbf{FFT}&\textbf{FT head} &\multicolumn{1}{c|}{\textbf{LoRA}} & \textbf{Prefix tuning} &
\textbf{Prompt tuning} & \textbf{P-tuning}   \\\midrule

 GPT2&117M& 0.6012 &  0.5362 &\multicolumn{1}{c|}{ 0.4830} & 0.4972 & 0.4869 & 0.4891 \\
 GPT2-medium&345M& 0.6965 &  0.5477 & \multicolumn{1}{c|}{0.4834} & 0.4893 & 0.4806 & 0.4930 \\
 GPT2-large&762M& 0.9037 &  0.5941 & \multicolumn{1}{c|}{0.4853} & 0.4959 & 0.4892 & 0.4845 \\
 GPT2-xl&1.5B& 0.9897 &  0.6276 & \multicolumn{1}{c|}{0.4898} & 0.4910 & 0.4916 & 0.4850 \\
 LLaMA3-1B&1B&  0.9999 &  0.9319 & \multicolumn{1}{c|}{0.4873} &  0.4981& 0.4849 & 0.4844\\
 LLaMA2-7B&7B& 0.9999 &  0.8886 & \multicolumn{1}{c|}{0.7865} &  0.4843&  0.4783 & 0.4871\\
\bottomrule
\end{tabular}
}

\caption{AUC scores of the LOSS attack across different fine-tuning methods and model sizes on the Wikitext dataset. All models are fine-tuned for 5 epochs using identical hyperparameter configurations.}
\label{tab:gpt_family_auc}

\end{table*}

Beyond tuning paradigms themselves, the scale of the underlying pre-trained model may also play a key role in determining memorization behavior. This raises an important question: \textit{To what extent does model size influence memorization under different fine-tuning strategies? }

\begin{tcolorbox}[sharp corners, colback=white!30, colframe=black, title=\textbf{Observation $\sharp$3}]
Model size significantly enhances memorization in parameter-based fine-tuning methods, while prompt-based methods show minimal sensitivity and maintain consistently low memorization.
\end{tcolorbox}

To systematically investigate the impact of different pre-trained model sizes on memorization across various fine-tuning methods, we conduct experiments across four variants of the GPT-2 architecture: GPT-2 (124M), GPT-2 Medium (345M), GPT-2 Large (762M), and GPT-2 XL (1.5B). To facilitate a fair comparison with LLaMA2-7B, we also include experiments on LLaMA3-1B, a smaller model in the same family. All models share the same underlying architecture and are pre-trained on the same dataset. This controlled setup allows us to isolate the effect of model scale while avoiding confounding factors such as architectural differences and variations in pre-training datasets. All models are fine-tuned using identical hyperparameter configurations for 5 epochs. 

The results presented in Table~\ref{tab:gpt_family_auc} reveal distinct trends between parameter-based fine-tuning methods (FFT, FT head, and LoRA) and prompt-based fine tuning methods (Prefix tuning, Prompt tuning, and P-tuning) across both datasets. Overall, as the size of the pre-trained model increases, parameter-based fine-tuning methods exhibit a pronounced increase in memorization capability, whereas prompt-based methods show a much weaker correlation with model scale.

Among the parameter-based methods, FFT achieves the highest memorization scores, showing a clear positive correlation with increasing model size. For instance, the AUC under FFT increases from 0.8892 with the smallest GPT-2 model to 0.9837 with GPT-2 XL, indicating that larger models are more susceptible to memorizing training data when all parameters are updated. FT head results in lower AUC scores than FFT, but still exhibits a positive correlation with model size.  Notably, LoRA exhibits relatively low memorization scores across all GPT-2 model sizes and LLaMA3-1B, which contrasts sharply with its strong memorization performance observed in the LLaMA2-7B experiments. We hypothesize that LoRA requires a larger parameter budget to effectively induce memorization in tasks such as Wikitext.

In contrast, the prompt-based tuning methods exhibit relatively stable yet significantly lower memorization scores regardless of model size. For example, the AUC for FFT rises from 0.6012 with the smallest GPT-2 model to 0.9897 with GPT-2 XL, and reaches nearly perfect 0.9999 on LLaMA3-1B. Meanwhile, prompt-based methods consistently score between 0.48 and 0.50 across all model sizes, showing no significant gains with larger models. This pattern further highlights the low sensitivity of prompt tuning to model scale, maintaining relatively stable yet limited memorization capability.

The findings demonstrate that parameter-based fine-tuning methods exhibit a pronounced increase in memorization as model size grows, likely attributable to the expanded parameter space allowing for more extensive adaptation to the training data—particularly in FFT, which updates all model parameters. LoRA, as a low-rank adaptation technique, shows enhanced memorization predominantly in larger models, indicating a dependency on sufficient parameter capacity to effectively capture memorization signals. Conversely, prompt-based tuning methods, which optimize only a limited set of input prompts rather than updating model parameters, maintain relatively stable but consistently lower memorization levels across model scales. This is because prompt tuning primarily influences model behavior through input modification without altering the underlying model weights or data distribution, thereby limiting the model’s ability to overfit or memorize fine-tuning data.

\begin{tcolorbox}[sharp corners, colback=white!30, colframe=black, title=\textbf{Observation $\sharp$4}]
Prompt-based tuning leads to stronger memorization in structured tasks than in other downstream tasks.
\end{tcolorbox}

\subsection{Impact of Downstream Tasks}\label{sec:6.2}
To investigate how downstream tasks influence model memorization under different fine-tuning strategies, we fine-tune LLaMA2-7B using various methods and evaluate the LOSS attack against the resulting models on three representative datasets: WebNLG, Wikitext, and Xsum. 

As shown in Table~\ref{tab:AUC}, prompt-based fine-tuning methods achieve significantly higher MIA AUC scores on the WebNLG dataset compared to other tasks. Specifically, while their AUC scores remain below 0.53 on Wikitext and Xsum, all prompt-based fine-tuning methods surpass 0.58 on WebNLG, with P-Tuning reaching as high as 0.6260. This performance gap may be attributed to the unique nature of the WebNLG task. In contrast to Wikitext’s language modeling and Xsum’s abstractive summarization, WebNLG involves structured text generation from triples—a task that demands strong semantic alignment between input entities and the resulting output text. Prompt-based methods, which steer model behavior through learnable discrete or continuous prompts, appear particularly well-suited to capturing such structured regularities. However, this adaptation may inadvertently reinforce memorization patterns in the generated text, as the task's deterministic input-output mappings intensify the model's reliance on memorized syntactic templates.

\subsection{Impact of LoRA Placement on Memorization}\label{sec:6.3}
LoRA has demonstrated a promising trade-off between model performance and privacy risk, exhibiting competitive performance while maintaining a lower risk of privacy leakage. Given these advantages, we further investigate the impact of LoRA’s configuration within transformer blocks on model memorization. Specifically, we conduct experiments placing LoRA adaptation matrices in different locations: the attention layer (LoRA\_attn) ,which is the default setting and assessed in sec~\ref{sec:5}, the projection layer (LoRA\_proj), and both layers simultaneously (LoRA\_attn\&proj), evaluating their effects on two different datasets with all the MIA methods.

Our results, summarized in Table~\ref{tab:lora_auc}, reveal that applying LoRA to the projection layer results in stronger memorization than applying it to the attention layer. Moreover, configuring LoRA in both layers leads to the highest memorization effect among the three settings. These findings align with previous studies~\cite{meng2022locating}, which have suggested that memorization in transformer-based models is primarily concentrated in projection layers. This can be attributed to the role of the projection layer in feature transformation and information compression, making it more susceptible to retain training data. Our results further reinforce this hypothesis, demonstrating that modifying the projection layer significantly increases the memorization tendency of the model, whereas altering only the attention layer has a more limited impact.
\begin{table}[h] %
\centering
\resizebox{0.5\textwidth}{!}{
\begin{tabular}{llllll}
\hline
\multicolumn{2}{l}{}           & LOSS & Ref & Zlib & Min-k\%  \\ \midrule
\multirow{3}{*}{wikitext}   & LoRA\_attn   &   0.5515    &     0.5712       &    0.5682     &    0.5425         \\ 
                            & LoRA\_proj   &   0.5788    &     0.6113       &    0.5951     &     0.5823          \\ 
                            & LoRA\_attn\&proj     &   0.6162  &    0.6644      &     0.6327    &    0.6224        \\ \midrule
\multirow{3}{*}{webnlg}     & LoRA\_attn   &    0.7270     &       0.6845       &     0.6952            &      0.5678         \\ 
                            & LoRA\_proj   &    0.7781     &       0.7123       &     0.7586            &      0.6061         \\ 
                            & LoRA\_attn\&proj     &    0.8037     &       0.7273       &      0.7886           &     0.6164      \\ \bottomrule
\end{tabular}
}
\caption{Attack performance (AUC) on LLaMA2-7B across different datasets and attack methods, with varying insertion positions of LoRA.}
\label{tab:lora_auc}
\end{table}

\section{Conclusion and Future Work}\label{sec:7}
In this study, we systematically reveal, for the first time, the differences in memorization of LLMs under different fine-tuning paradigms. Our experimental results demonstrate significant variations in the extent to which different fine-tuning methods retain fine-tuning data, with prompt-based fine-tuning exhibiting substantially lower memorization, particularly in large-scale models. This finding indicates that parameter-based fine-tuning poses a considerably higher memorization risk compared to prompt-based fine-tuning, highlighting the importance of selecting appropriate fine-tuning strategies for privacy-sensitive applications.

While this study provides a foundational analysis of memorization across different fine-tuning paradigms, several important directions remain open for further exploration. Future research can extend our findings to larger-scale models (e.g., GPT-4, LLaMA2-70B) and more diverse architectures, such as multimodal models, to assess the generalizability of our conclusions across different scenarios. Additionally, a more comprehensive examination of advanced privacy attacks, including data extraction attacks and attribute inference attacks, could provide deeper insights into the interplay between model memorization and privacy risks from multiple perspectives.
\section*{Limitations}
Our study evaluates the memorization effects of different fine-tuning methods using MIAs on several language models, including the GPT-2 series, LLaMA2-7B, and LLaMA3-1B. While these models span medium-scale and large-scale architectures, our findings may not fully generalize to even larger models or other architectures such as mixture-of-experts (MoE) models. Additionally, our experiments are conducted on a limited set of datasets, which may not capture the full variability of real-world training distributions. Nonetheless, the consistent patterns observed across models and datasets provide strong empirical evidence of the disparity in memorization between parameter-based and prompt-based fine-tuning. These findings underscore the importance of selecting fine-tuning strategies carefully, particularly in privacy-sensitive applications.

\section*{Ethics Statement}

In this study, we use publicly available data sets and open source models, ensuring that our work does not involve any concerns about privacy or copyright. Our research aims to investigate the extent of data memorization across different fine-tuning paradigms, ultimately providing valuable insights to guide users in selecting appropriate fine-tuning strategies for LLMs.

\newpage
\bibliography{acl_latex}

\end{document}